\documentclass[letterpaper, 10 pt, conference]{ieeeconf}  

\IEEEoverridecommandlockouts                              

\usepackage{float}
\usepackage{subfigure}
\usepackage{multirow}
\usepackage{graphicx} 
\usepackage{amsmath} 
\usepackage{hyperref}

\newtheorem{defn}{Definition}[section]

\title{\LARGE \bf
	Topological Area Graph Generation and its\\ Application to Path Planning
}

\author{Jiawei Hou, Yijun Yuan and S\"oren Schwertfeger 
	\thanks{The authors are with the School of Information Science and Technology, ShanghaiTech University,
			Shanghai 201210, China, 
	and also with the University of Chinese Academy of Sciences,
			Beijing 100049, China 
			{\tt\small [houjw, yuanwj, soerensch]@shanghaitech.edu.cn}}
}

\begin{document}

	\maketitle
	\thispagestyle{empty}
	\pagestyle{empty}

	\begin{abstract}
%

    Representing a scanned map of the real environment as a topological structure is an important research in robotics. 
    Since topological representations of maps save a huge amount of map storage space and online computing time, they are widely used in fields such as path planning, map matching, and semantic mapping.
    
	We propose a novel topological map representation, the Area Graph, in which the vertices represent areas and edges represent passages. The Area Graph is developed from a pruned Voronoi Graph, the Topology Graph. The paper also presents path planning as one application for the Area Graph. For that, we derive a so-called Passage Graph from the Area Graph.
	Because our algorithm segments the map as a set of areas, the first experiment compares the results of the Area Graph with that of state-of-the-art segmentation approaches, which proved that our method effectively prevented over-segmentation. 
	Then the second experiment shows the superiority of our method over the traditional A* planning algorithm.

	\end{abstract}

\section{INTRODUCTION}
	Robotics has seen tremendous developments in recent years. There are more and more mobile autonomous robots that are active in bigger and more complex areas for long times. This poses challenges for the storage and application of traditionally used 2D grid maps - they grow too big and planning on them will be infeasible for very big areas. 
	The obvious solution to this problem is to use topological map representations, a solution already used by car navigation systems. 
	
	Rather than relying on provided maps, an important aspect of autonomous robotics is the generation of the maps using the robots sensor data. This mapping process, which is often embedded in a Simultaneous Localization and Mapping (SLAM) algorithm, ensures that the robot can use up-to-date maps and react to changes in the environment, e.g. to re-plan a path if a certain route is blocked. 
	
	The canonical output of most SLAM algorithms is a 2D grid map, from which a topological map should then be generated. 
	In our work, we present a method to extract a novel topological representation, the Area Graph, from a 2D grid map. We extract areas in the environment based on the Topology Graph presented in \cite{Schwertfeger2013TopoICRA} and \cite{Schwertfeger2016PathMatching}. From those areas, a graph where the vertices represent the areas and the edges represent the common boundaries between two areas, i.e \emph{passages}, is created as the Area Graph. 
We believe that our representation is more useful than classical topological representations, because represent areas instead of places, which allows a more intuitive usage of topological maps.
As an example use case, we use the Area Graph to perform path planning.
	
	The main contributions of our paper are the definition of the Area Graph and the algorithm for automatically generating the graph from a 2D grid map. Furthermore, we present how we can then use a data structure derived from the Area Graph, the \emph{Passage Graph}, 
	for path planning. 

	The paper is structured as follows. In Section~\ref{sec:generation_of_areagraph}, 
	the Topology Graph presented in \cite{Schwertfeger2013TopoICRA} and \cite{Schwertfeger2016PathMatching} is briefly described
	and the approach for the Area Graph generation is explained in detail.
	Since the Area Graph segments the environment into areas, we present, in Section~\ref{sec:comparison_segmentation}, related works on map segmentation. We subsequently compare the results of our algorithm with state of art segmentation algorithms. In Section~\ref{sec:application_planning}, the planning application based on the Area Graph is implemented and the performances of our method with two different types of roadmaps are compared with planning using A*. 
	Conclusions are drawn in Section~\ref{sec:conclusions}.
	

	
\section{Generation of the Area Graph}
\label{sec:generation_of_areagraph}
	In this section, a brief overview of the steps for generating a Topology Graph \cite{Schwertfeger2013TopoICRA} is recalled first. 
	Then, a detailed description of the algorithm which generates areas for the Area Graph is presented. The method is twofold: first, areas are generated from the Topology Graph, then areas in the same room are merged together.
	
	\subsection{Topology Graph Generation}
	\label{sec:topo_generation}
	
	The Topology Graph described in \cite{Schwertfeger2013TopoICRA}, can be defined as follows:
	\begin{defn}
		A Topology Graph is a topological structure $ G_T=(V_T,E_T) $ with a set of vertices $ V_T=\{v_0, \dots , v_{n} \} $ and a set of edges $ E_T=\{ e_0, \dots, e_{m} \} $， which is derived from the Voronoi Diagram (VD). 
		
		An edge $ e_k=(v_i,v_j) ~ ( k\in[0,m], ~ i,j\in [0,n], ~ i \neq j ) $ is a polyline connecting the \emph{waypoints}, i.e. a path which is formed by an ordered list of \emph{waypoints} from $ v_i $ to $ v_j $. The waypoints are vertices of the initial Voronoi Graph computed from a grid map.
	\end{defn} 
	

	Because the map given by a robot includes some noise data in the unmapped area, 
	the boundary of the map needs to be found to remove the redundant data. 
	For that purpose the alpha shape algorithm \cite{Edelsbrunner1983On} is used. 
	The algorithm uses the CGAL 2D Alpha Shapes \cite{cgal:d-as2-17b} to generate alpha shapes, and the biggest alpha shape is regarded as the boundary. All the vertices and edges outside such boundary are filtered out.
	
	Since the unpruned VD from the 2D point map has edges going between two close obstacles with a small space, i.e. points at walls, all edges whose distances to occupied cells that are smaller than a threshold are deleted from the graph.
	This processing makes only dead-ends and junctions are vertices of the graph. During this step, the edges of a two-degree vertex are joined as one edge. These two steps are repeated several times to make the graph simple enough, where the number of repetitions depends on some user-defined parameters. Those parameters are described in more details in the work of \cite{Schwertfeger2013TopoICRA}.
	
\begin{figure*}[t]
			\centering
			\subfigure[]{
			\label{fig:topoGraph}
				\includegraphics[width=0.19\linewidth]{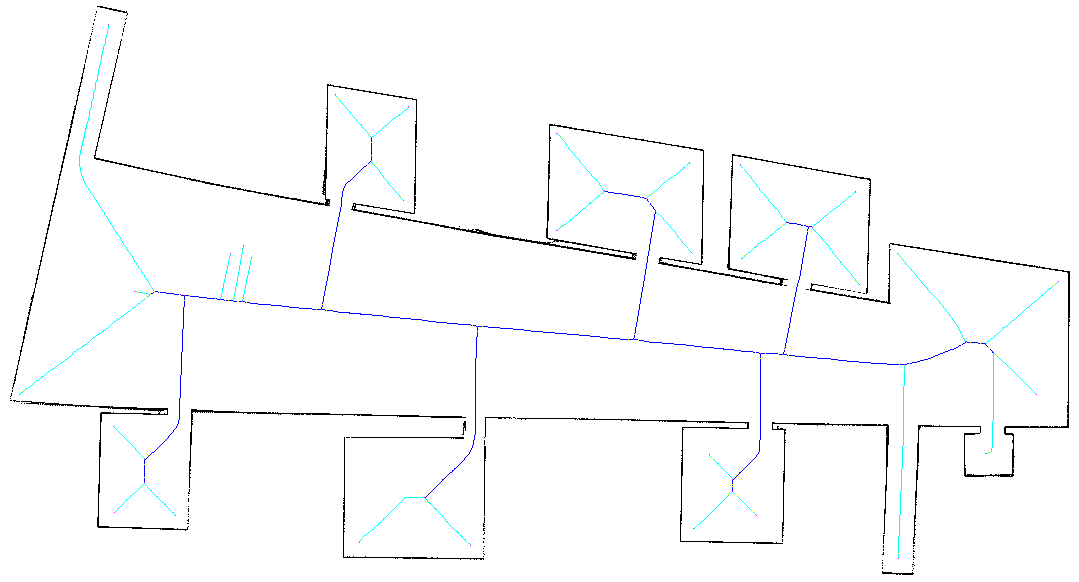}}
			\label{fig:lab2}
			\subfigure[]{
				\label{fig:alphaLab}
				\includegraphics[width=0.19\linewidth]{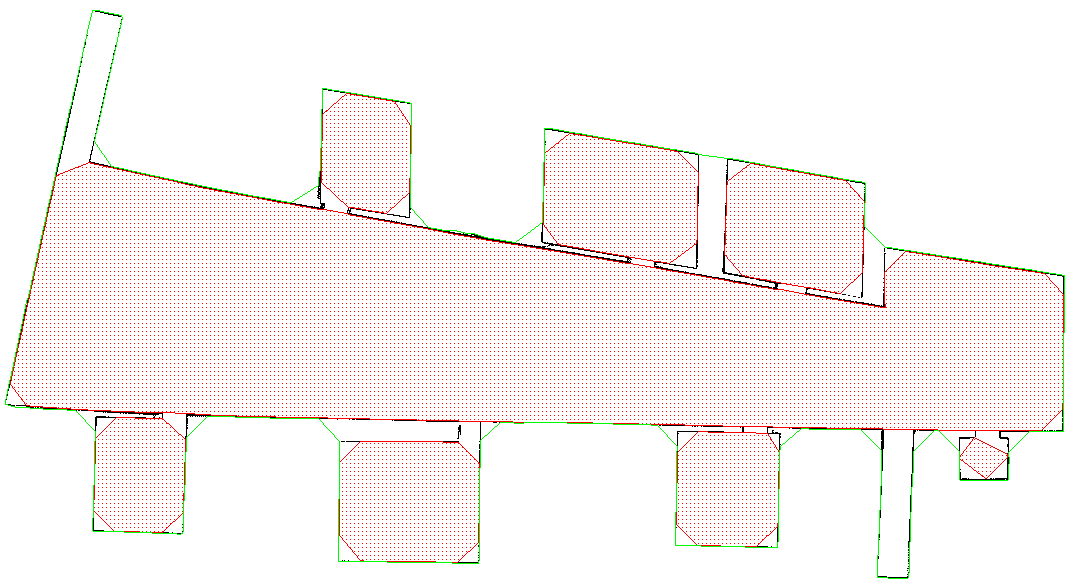}}
			\subfigure[]{
				\label{fig:lastVoriLab}
				\includegraphics[width=0.19\linewidth]{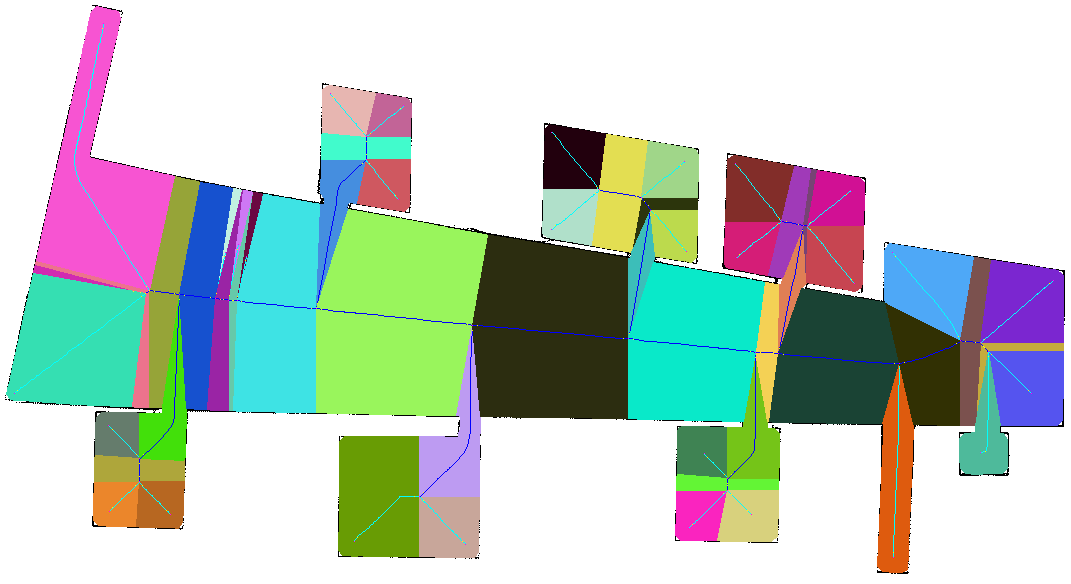}}
			\subfigure[]{
				\label{fig:dectRoomLab}
				\includegraphics[width=0.19\linewidth]{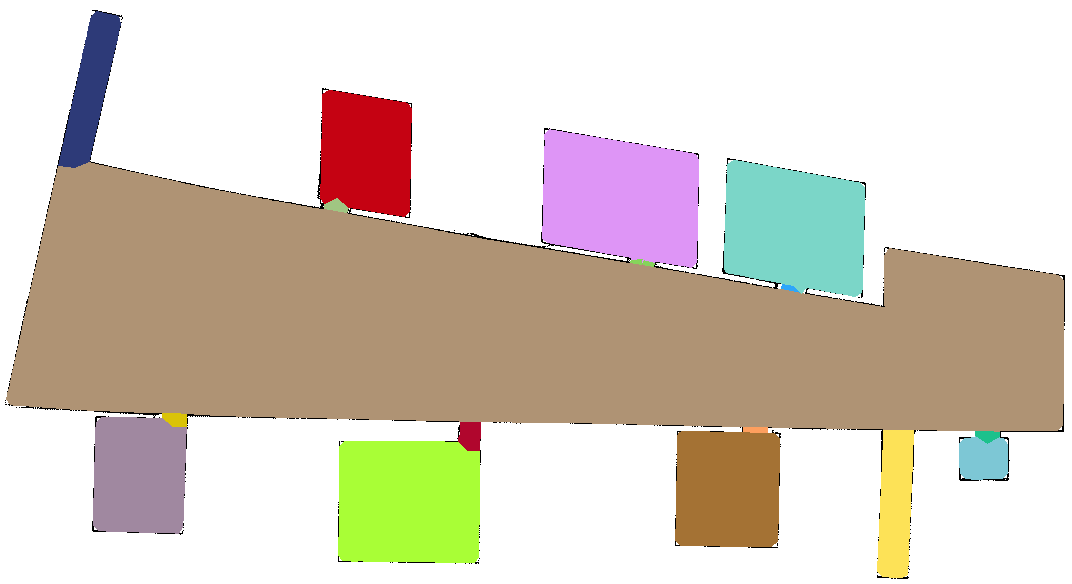}}
			\subfigure[]{
				\label{fig:passageGraph}
				\includegraphics[width=0.18\linewidth]{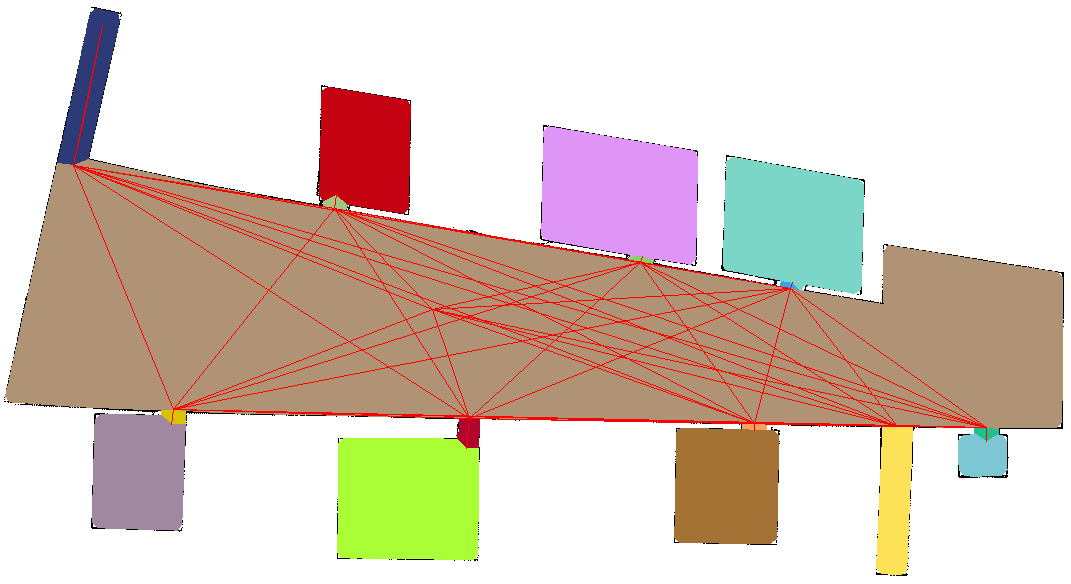}}
			\caption{
			(a) A Topology Graph obtained by pruning the Voronoi diagram using the method in \cite{Schwertfeger2016PathMatching}. Dead-end edges are shown in light-blue and edges connecting non-dead-end vertices (junctions) are shown in dark blue. The grid map (black) is from \cite{bormann2016room}.
			(b) Alpha shape detection to get the boundary (green line) and rooms (red pattern) from the grid map (black).
			(c) The areas before merged with the Topology Graph.
			(d) Areas within the same room (see (a)) have been merged here.
			(e) Map with areas and the edges (in red) of the Passage Graph. One can see how all of the passages of the big brown center area are connected to each other. Each of the edges is attributed with a path and the length of the path. The paths for the edges are not shown here.}
		\end{figure*}

	Since only the graph of reachable areas is useful, the connected sub-graph with the largest sum of length is kept and all vertices and edges not belonging to the biggest connected graph are removed. Finally, vertices that are too close are merged.
	Fig. \ref{fig:topoGraph} shows a Topology Graph constructed from a grid map.

	\subsection{Generating Areas from a Topology Graph}
	To generate the polygons for edges as areas, a graph derived from the Topology Graph is constructed first.
	The area generation steps are synchronized with the Topology Graph creation.
	We identify the notations for the graph derived from the Topology Graph as $ G_A=(V_A,E_A,P_A) $, whose vertices $ V_A $ and edges $ E_A $ are same as the vertices and edges in the Topology Graph, and polygons $ P_A $ are attached to edges as areas. 
	
		\begin{figure}[tpb]
			\begin{center}
				\includegraphics[width=1.0\linewidth]{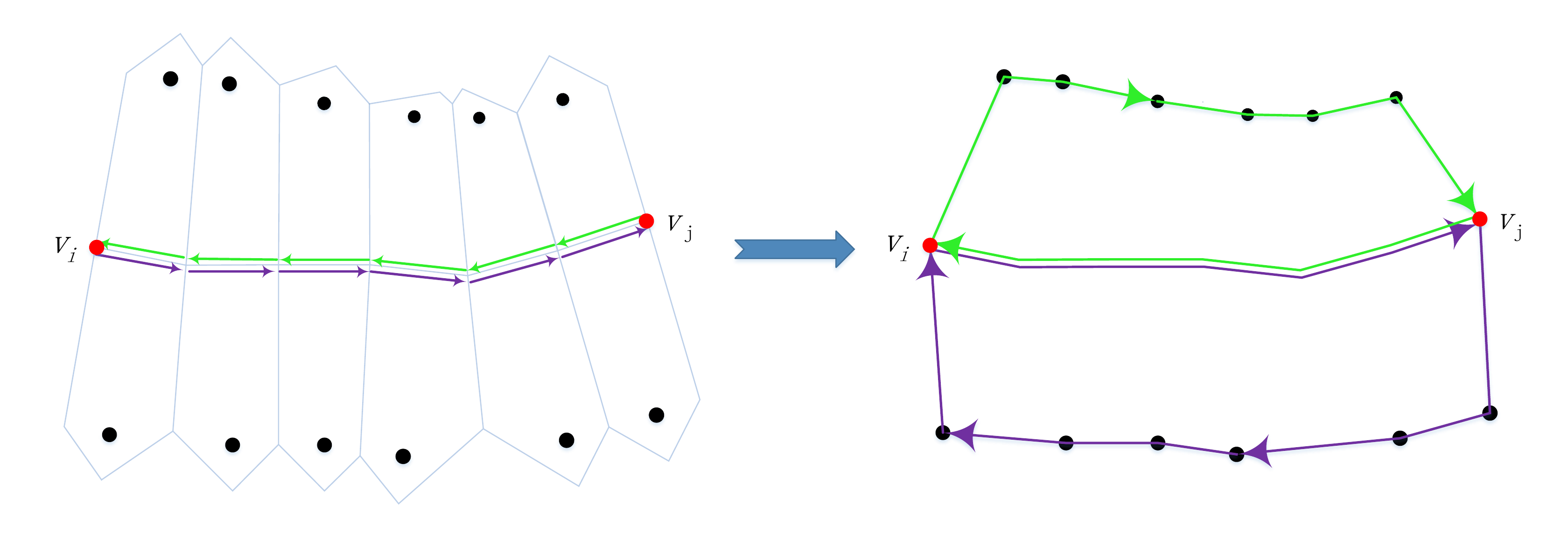}
			\end{center}
			\caption{Left schematic: The green short arrows are the Voronoi edges that make up the halfedge $ h_{ji} $. Correspondingly, the purple short arrows make up the half edge $ h_{ij} $ in posite direction. Each Voronoi edge has a face, which contains exactly one site inside. Right schematic: Connect two halfedges with sites in clockwise order respectively to create two half-polygons for an edge.} 
			\label{fig:faces}
		\end{figure}
		\begin{figure}[tpb]
		
			\subfigure[Join a polygon of a dead end edge to its next half-polygon, in the case that its next neighbor edge is not a dead-end.]{
				\label{fig:rightdeadend}
				\includegraphics[width=0.95\linewidth]{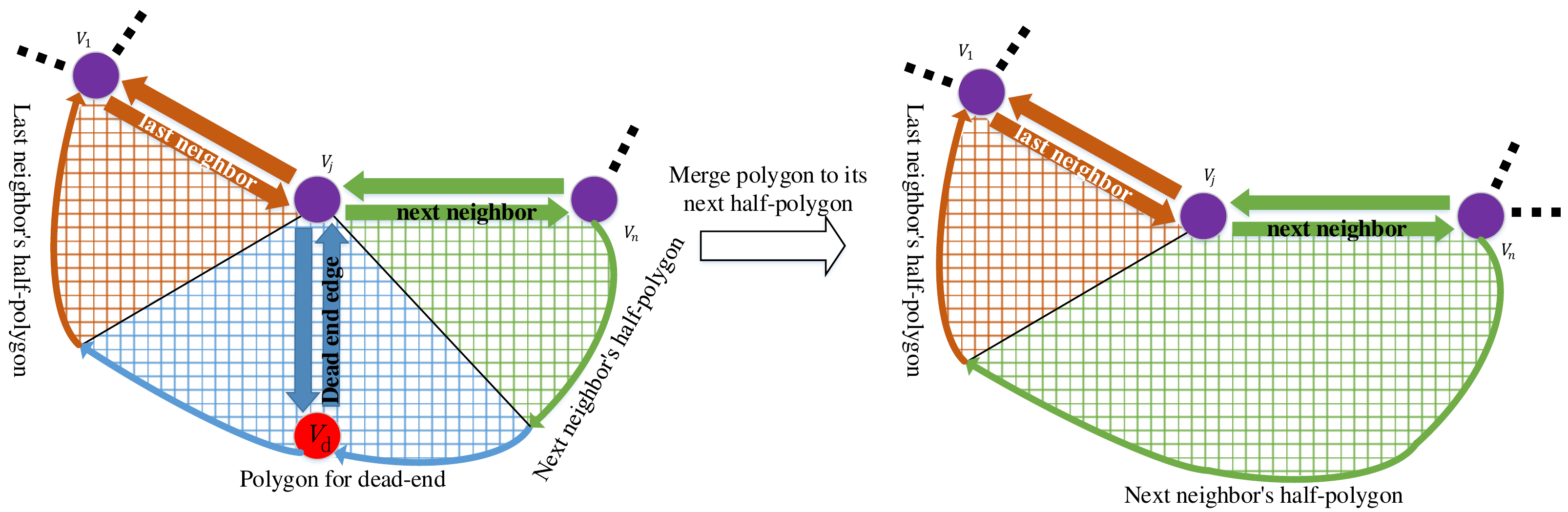}}
			\subfigure[Join a polygon of a dead end edge to its last  half-polygon, because the right neighbor of $h_{dj}$ is a dead-end.]{
				\label{fig:leftdeadend}
				\includegraphics[width=0.95\linewidth]{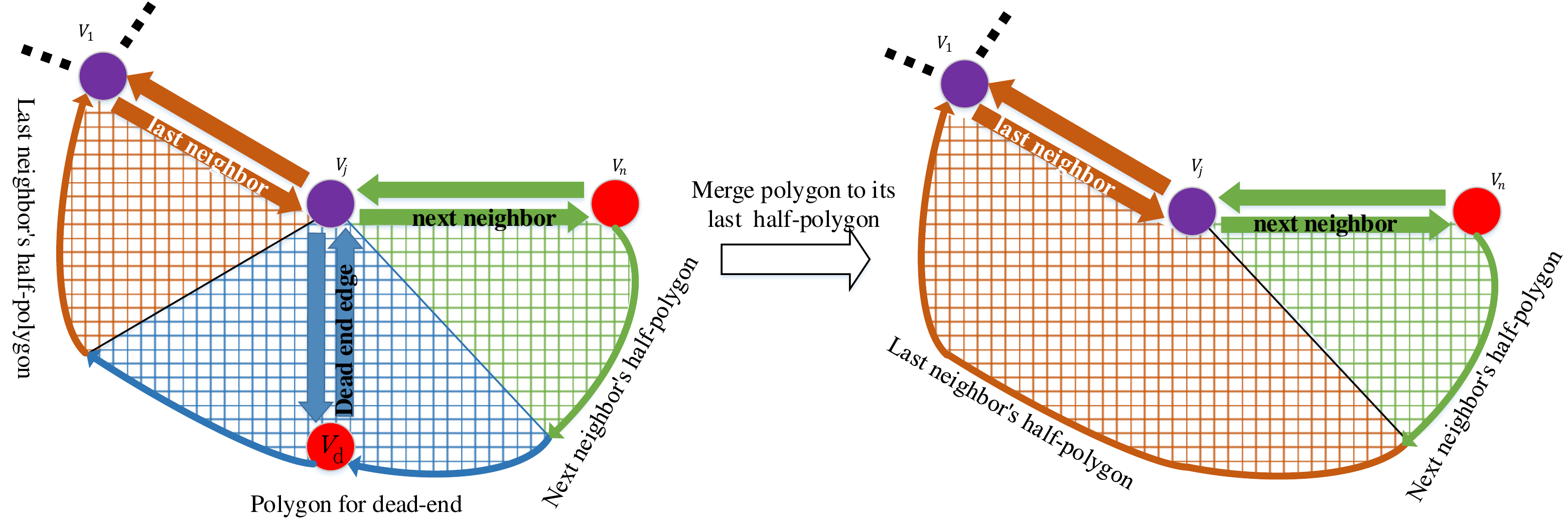}}
			\caption{The two cases of merging polygon of the dead-end edge into its neighbor's half-polygon (here only shows half polygons of $e_{dj}$'s neighbors instead of the whole polygon). Red points represent dead-ends, the edge connected to a dead-end is a dead-end edge. Purple points represent junctions. \label{fig:deadend_face}}
		\end{figure}

	First, the Voronoi Diagram is generated with the Computational Geometry Algorithms Library (CGAL) \cite{cgal:k-vda2-17b}. Then the VD is pruned and filtered as described in \cite{Schwertfeger2013TopoICRA} to produce a basic skeleton. An edge $ e_{ij}=(v_i, v_j) $ consists of two twin halfedges $ \{h_{ij}, h_{ji} \}  $ that are in opposite direction.  During the VD generation \cite{cgal:k-vda2-17b}, the faces for each of the halfedges are already saved. As a result, faces can be utilized to create a polygon for each edge. 
	For each halfedge $ h_{ij} $, we build a half-polygon $ hpoly(h_{ij}) $ by connecting the waypoints and the dual sites in the faces in clockwise order, as shown in Fig. \ref{fig:faces}. A pair of twin half polygons $ hpoly(h_{ij})$ and $hpoly(h_{ji}) $ are regarded as the polygon attached to the edge $ e_{ij} $, denoted as $ p_{ij}=poly(e_{ij})=\{hpoly(h_{ij}), ~ hpoly(h_{ji})\} \in P_A$.

	The next two steps are the same as the steps mentioned in the Topology Graph generation: deleting the part of the graph outside the boundary and filtering out edges that are shorter than a user-defined threshold. 
	The difference to the steps in the Topology Graph generation is that the polygons attached to the edges need to be taken into account.
	
	When the edges are joined, the polygons attached to the edges need to be merged. We merge the half-polygons of two joined halfedges by connecting the waypoints of the two halfedges and their sites in clockwise order. Then the two merged half-polygons make up the polygon of the edge. 
	
	
	When a dead-end edge is removed, the polygon attached to that edge needs to be merged into its neighboring polygons. Otherwise, the areas of the dead-ends would be lost.
	Fig. \ref{fig:deadend_face} is a schematic illustration of the process used for merging the polygons of dead-ends.
	If a  dead-end  edge  $ e_{dj}=\{h_{dj},h_{jd} \} $ needs to be deleted,  
	its polygon $ p_{dj}=poly(e_{dj}) $ will be merged into one of its two neighboring half-polygons. 
	Here we label the halfedge $ h_{jn} $ after $ h_{dj} $ in clockwise order as the \emph{next halfedge} of $ h_{dj} $, and the halfedge $ h_{lj} $ before $ h_{jd} $ in clockwise order as the \emph{last halfedge} of $ h_{dj} $. The terms \emph{last half-polygon} and \emph{next half-polygon} are defined in the same way.
	We prioritize merging the dead-end edge's polygon $poly(e_{dj})$ to its next half-polygon $ hpoly(h_{jn}) $. If $ h_{jn} $ is also a dead-end halfedge, then $poly(e_{dj})$ will be merged into its last half-polygon $ hpoly(h_{lj}) $. This strategy helps to avoid the imbalance of the polygon area for dead-end edges.
	Fig. \ref{fig:rightdeadend} shows both cases of the processing of merging the polygon of dead-end.
	Since the algorithm in \cite{Schwertfeger2013TopoICRA} removes dead-ends for several times, the steps merging polygons will also be performed for several times.

	\subsection{Merging Areas in the Same Room}
	 \label{sec:mer_room}
	
	Because it is not necessary to segment a room into different areas,
	the polygons generated in the previous section are not the final areas of the Area Graph. Fig. \ref{fig:lastVoriLab} is an example that shows polygons generated from the Topology Graph that do not correctly represent rooms.  
    Hence, we merge all the polygons that are in the same room. 
	
	To find the boundary of the mapped area, CGAL 2D Alpha Shapes \cite{cgal:d-as2-17b} is used to create Alpha Shapes. Different from the work in \cite{Schwertfeger2013TopoICRA}, our algorithm  saves all alpha shapes other than the largest one for detecting rooms.
	
	In Fig. \ref{fig:alphaLab}, we show an example of detected alpha shapes extracted from a map.  
	The open space inside the boundary detected by $ \alpha $-shape is regarded as a \emph{room}. 
	The minimum area of the room that can be detected depends on the parameter $ \alpha $.
	Note that the parameter $\alpha$ can be treated as the square of the radius of the largest empty disk that can be put into the detected rooms. 
	That means, $\alpha = R^2$, where $ R $ is the radius of the disk \cite{cgal:d-as2-17b}.
	If this disk can be put into an open space completely, then an $ \alpha $-shape can be detected in this space. 
	That is, in a detected room, there is at least one point whose square of the distance to its closest obstacle is larger than $ \alpha $.
	The larger $ \alpha $, the larger the smallest $ \alpha $-shape is. Hence, fewer rooms can be detected with a larger $\alpha$, and fewer polygons are merged. 
	
\begin{figure*}[t]
			\subfigure{
				\label{fig:lab_mlcolm}	\includegraphics[width=0.31\linewidth]{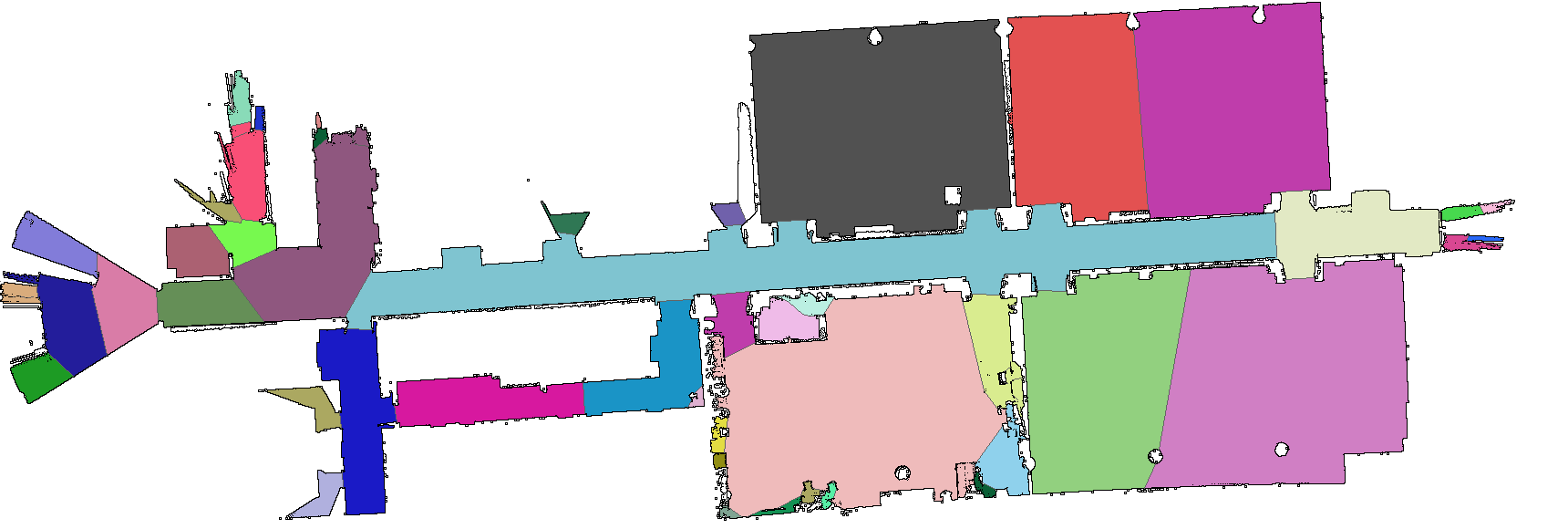}}
			\subfigure{
				\label{fig:lab_ipa}
				\includegraphics[width=0.31\linewidth]{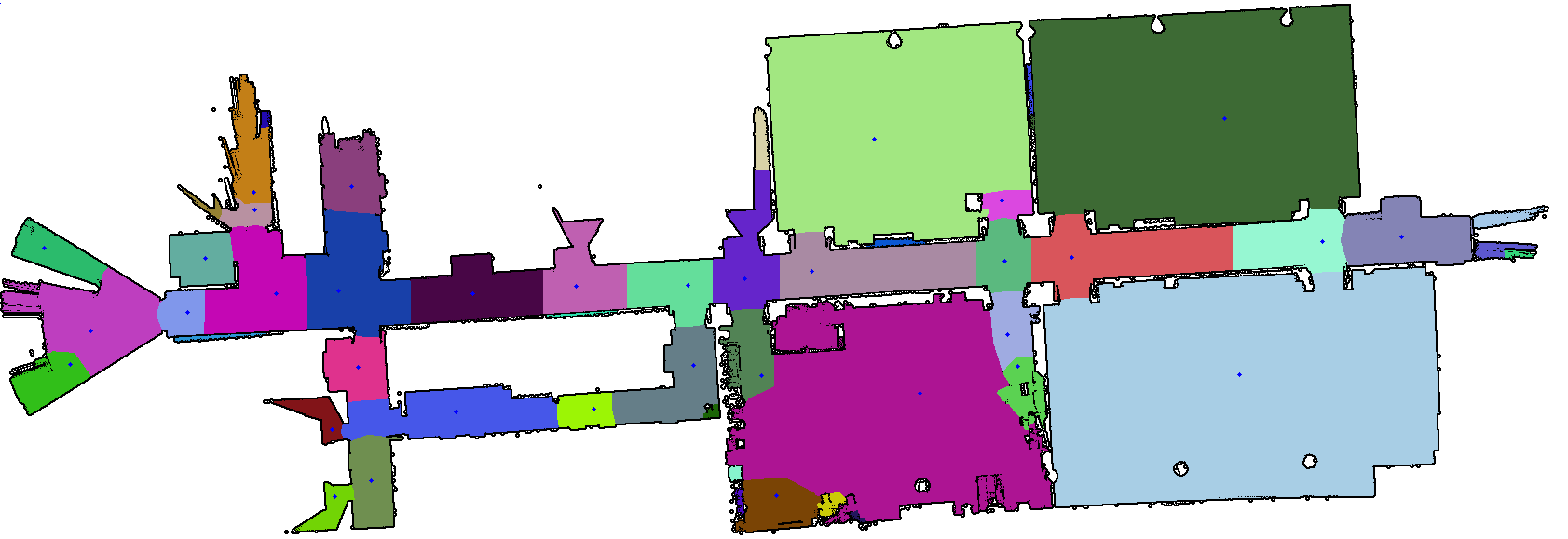}}
			\subfigure{
				\label{fig:lab_my}
				\includegraphics[width=0.31\linewidth]{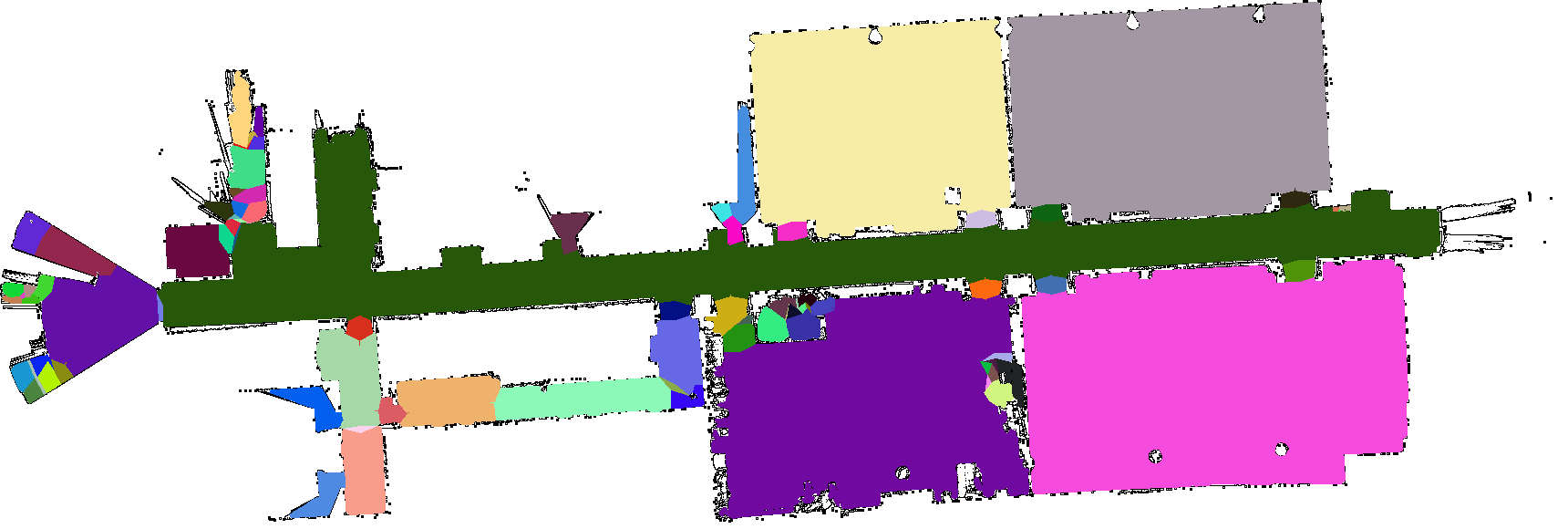}}
			\subfigure[MAORIS.]{
				\label{fig:d_mlcolm}
				\includegraphics[width=0.31\linewidth]{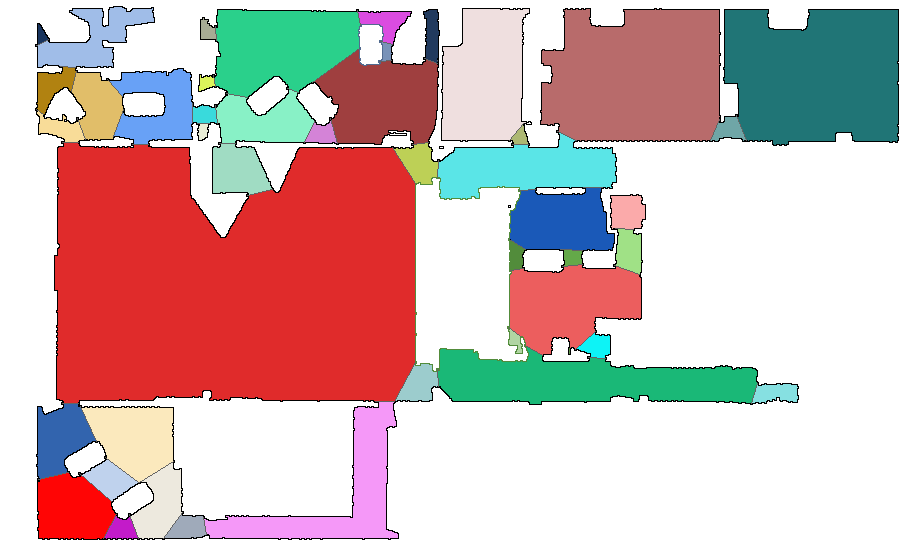}}
			\subfigure[Voronoi-based segmentation.]{
				\label{fig:d_ipa}
				\includegraphics[width=0.31\linewidth]{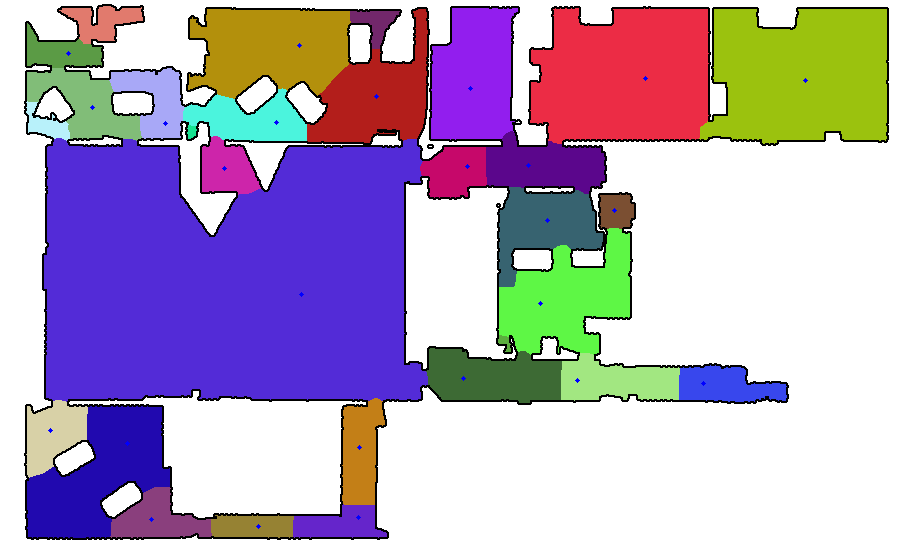}}
			\subfigure[Our method.]{
				\label{fig:d_my}
				\includegraphics[width=0.31\linewidth]{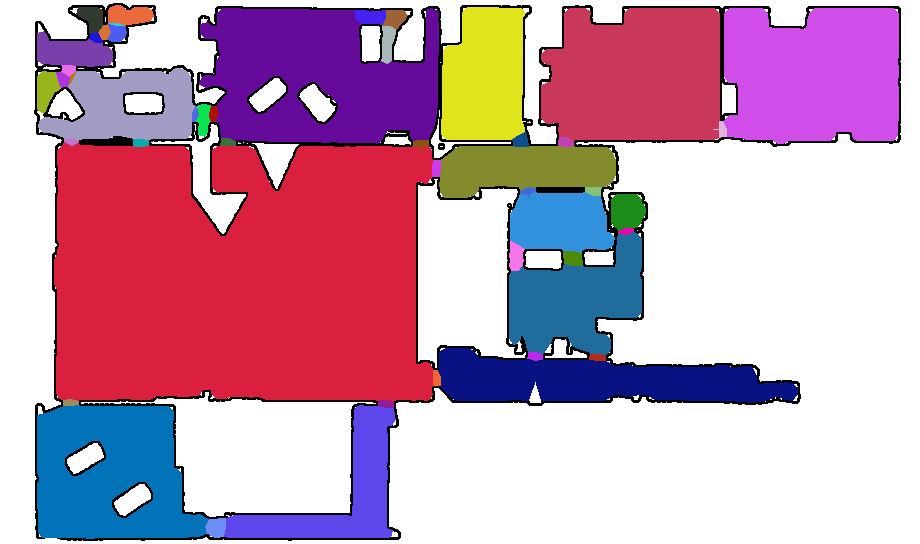}}
			\caption{Compare the segmentation results from three different methods.\label{fig:lab_seg}}
		\end{figure*}

	The $ \alpha $ value is decided according to the resolution of the map, and the widths of doors and corridors in the environment. In our implementation for the Robot Operating System (ROS), the resolution can be obtained from the yaml file automatically. 
	Taking the fist map shown in Fig. \ref{fig:lab_seg} as an example,
	the map's resolution is 0.05, the width of the widest door is 1.64 meter and the width of the narrower part of the corridor is 2.42 meter.  The calculation of the $ \alpha $ based on the environment is demonstrated here. 
	
	First, convert the unit of the width from meters to pixels with
	\begin{equation}
		\text{Width in Pixels}= \frac{\text{Width in Meters} } {\text{Resolution}}. 
	\end{equation}
	To detect rooms, the most important point is to ensure that the virtual disk cannot pass through the door. Thus, the radius  of the disk cannot be smaller than half the width of the door $ W_d $. Thus $ \alpha $ should satisfy:
	\begin{equation}
		\alpha = R^2 \ge (\frac{W_d}{2})^2
	\end{equation}
	In our example map, the condition is computed as such:
	\begin{align*}
		 W_d= \frac{1.64}{0.05} = 32.8, \qquad
		 \alpha \ge (\frac{32.8}{2})^2 = 268.96
	\end{align*}
		
	If we want to segment the whole corridor as one area, 
	the virtual disk should be put into the corridor entirely. Hence, the disk's radius $ R $ should not be larger than half of the width $ W_c $ of the narrower part of the corridor: 
	\begin{equation}
		\alpha = R^2 \le (\frac{W_c}{2})^2
	\end{equation}
	In our example map, the condition is:
	\begin{align*}
		 W_c = \frac{2.42}{0.05} = 48.4, \qquad
		 \alpha \le (\frac{48.4}{2})^2 =585.64
	\end{align*}
	
	Hence, for this example map, to segment a corridor as one area, the $ \alpha $ value should satisfy $ 268.96 \le \alpha \le 585.64 $. 
	
    Besides, one should note that the $ R $ can not be larger than half the width of the smallest room that we want to detect, otherwise, some small rooms cannot be detected.

	After obtaining the $ \alpha $-shape to find rooms, polygons in the same room need to be merged as an area to represent a complete room. For this purpose, we split the polygons crossing the boundary between rooms, and merge the polygons belonging to the same room.
	
	We determine if a polygon belongs to a room by judging whether the edge that it belongs to is inside the $ \alpha $-shape. If both endpoints of the edge are inside the $ \alpha $-shape or only one endpoint of the edge is inside the $ \alpha $-shape but the edge is a dead-end, this edge and its polygon are judged as belonging to the room detected by this $ \alpha $-shape. 
	
	If an edge crosses the $ \alpha $-shape, i.e. one of its endpoints is inside and the other endpoint is outside the $ \alpha $-shape and the edge is not a dead-end, its polygon is divided into two at the \emph{passage line}.
	Here, a passage line is generated by connecting the intersection of the $ \alpha $-shape and the edge to its two closest sites on its two sides respectively. 
	
	All polygons belonging to the same room are assigned the same $ roomID $. Then we merge the polygons with the same $ roomID $ in one polygon, representing a room. Fig. \ref{fig:dectRoomLab} shows the area after merging polygons.

	Through the above steps, an Area Graph can be created. The areas are regarded as vertices of the Area Graph. The neighbors for each area are recorded and the passages between areas are the edges of the Area Graph connecting adjacent areas.

\section{Comparing with Works on Segmentation}
\label{sec:comparison_segmentation}

	
	The comparison between the Area Graph and other methods in segmentation is shown with a discussion and an experiment.
	
	\subsection{Compare with State-of-the-art Segmentation}
	
	Fundamentally, a basic contribution of Area Graph is that it represents a map as a set of areas, which can be regarded as a segmentation for a map. 
	
	Bormann et al. \cite{bormann2016room} introduced various kinds of methods on room segmentation and compared the segmentation results of those methods. 
	And the comparison experiment of \cite{bormann2016room} shows that Voronoi-based segmentation results have the highest degree of approximation of the ground truth.

	In \cite{Thrun1998Learning}, a Voronoi graph-based segmentation method is shown, which creates a topological representation using region cells that represent rooms or parts of rooms by means of critical points. \emph{Critical point} is the point lying closer to obstacle points than all its neighboring points on the Voronoi graph. Thus critical points usually locate at narrow passages such as doorways. 
	After that, the Voronoi-based segmentation works \cite{Kai2008Coordinated} with some optimization to select the critical points only at real doors is proposed.
	To segment complete rooms, some heuristics are used to merge small segments \cite{bormann2016room}. 

	 Mielle et al. \cite{Malcolm2017MAORISICRA} segment maps by calculating distance image from them, in which each pixel has a pixel value that represents the size of the region it belongs to, i.e. the distance to its closest obstacle.  Then they 
	 merge regions with similar values. This method helps to relieve over-segmentation on corridors. 
	 Mielle et al. have published their code online. We will compare our segmentation results with theirs.

	 Compared to other categories of methods, our method has two main advantages. 
	 First, the positions of passages are stored when the areas are generated, which makes it easier for us to do path planning on the area representation, especially compared with \cite{Thrun1998Learning}.
	 Second, most methods tend to segment the whole corridor as an area, although our algorithm also good at doing this while distinguishing a corridor into straight-walk parts and the junctions can be a useful and interesting task. Our algorithm can achieve this requirement by only changing a parameter,  $\alpha$.

	Compared to Voronoi-based segmentation, our method looks for open space by detecting alpha shapes as rooms instead of finding critical points and setting heuristics, which makes our method simpler and avoid the trouble for adjusting parameters in heuristics.

	\subsection{Comparison Experiments}
	\label{experiments_seg}
	
	This experiment compares our algorithm with Mielle's MAORIS methods \cite{Malcolm2017MAORISICRA} and the Voronoi-based segmentation from Bormann's implementation \cite{bormann2016room}. 	Because of the reason that how the developer would like to segment a map is very subjective, we just analyze the segmentation results, without  comparing them to subjective ground truths. Due to limited space, we only show the segmentation results of the three methods for the two maps in Fig. \ref{fig:lab_seg}, but you can get more segmentation results by running our code that has been provided online\footnote{\url{AvailableInTheFinalVersion}} on more maps.

	The relationship between the widths of the doors and corridors in the environment has been analyzed in Section~\ref{sec:mer_room}.
	The first map that we use in Fig. \ref{fig:lab_seg} is a map scanned from a real environment and edited to remove some noise points and add missing walls.
	We choose $ \alpha = 420 $ when running our algorithm on the first map. The second map is from Bormann's dataset \cite{bormann2016room}. We measure this map with a graphical tool. The width of doorways in this map is 14 pixels and  its narrower corridors (at the bottom left of the map) is about 20 pixels wide. Therefore, in order to segment the entire corridor into one area, $\alpha$ needs to satisfy $49\le \alpha \le 100$. Then we choose $ \alpha = 81 $ for this map.
	
	As can be seen from the Fig. \ref{fig:lab_seg}, all three methods can separated rooms and corridors perfectly. Now that all these methods have achieved this basic goal of segmentation, we will discuss the differences between them of segmentation in rooms and corridors.
	
	The obvious conclusion from the first row of the Fig. \ref{fig:lab_seg} is that our algorithm performs better to segment a long corridor as a complete area, while the other two methods over-segment the corridor. 
	It can be observed that, in the bottom left room of the first map, the three methods are all affected by the furniture in the room, which leads to some over-segmentation. Since the MAORIS method is based on distance to obstacles, it tends to connect the protrusions obstacles from the walls. Therefore, the presence of a pole in an empty room can lead to over-segmentation for this method. Owing to the use of the merging heuristics in Voronoi-based segmentation, this method divides only one more area at the place where some furniture are, while our method produces a few more regions there. However, since these areas are too small and can be dynamically changed, they are less likely to be used, for example, for location and navigation.
	
	The major feature of the map in the second row of Fig. \ref{fig:lab_seg} is that there are many irregular holes in the rooms. In this case, our method still shows to be accurate in identifying a complete room.
	
	In summary, our approach performed outstandingly in avoiding over-segmentation, which is mainly due to the room detection and merging algorithm.
	
\section{Application to Path Planning}
\label{sec:application_planning}
	
	Segmenting grid maps into appropriate regions is an important task for many applications in robotics. Planning, one of its fundamental application, is implemented in this work, for which we create a roadmap graph, the Passage Graph, based on the Area Graph.
	

	\subsection{Introduction of Graph-based Planning}
	\label{subsec:compare_planning}
	Graph-based planning, also called roadmap-based planning, constructs a one-dimensional graph from a 2D grid map to save paths between vertices off-line, then a path can be extracted from this graph rapidly with simple search in on-line query stage.
	Roadmap-based methods can be distinguished according to the variety of their underlying approach, such as Visibility graphs \cite{Latombe1991Robot}, Probabilistic Roadmap Methods (PRM) \cite{berg2005using,kavraki1996probabilistic,holleman2000a} and Rapidly-exploring Random Trees (RRT) \cite{Kuffner2002RRT,Martin2007Offline,Yang2011Anytime}.
	

    Some planning algorithms  \cite{hoff2000fast,rohnert1988discs} use the Voronoi diagram for obtaining paths which have the maximum amount of clearance from obstacles to themselves. 
    To build a Voronoi-based roadmap, our implementation has saved the pruned Voronoi diagram, the Topology Graph, as the walkable paths, which saves the time of forming paths.
	Compared with these Voronoi-based planning, our roadmap has fewer vertices thanks to the room merging step, which leads to an effective reduction of the number of passage vertices.

    In some other application scenarios, such as simulations and virtual environment of games, the navigation mesh is widely employed in path planning for characters. The navigation mesh methods \cite{geraerts2010planning,Van2016A,hale2008automatically} partition the walkable environment into a set of 2D regions, then a virtual character can choose its movements inside each region \cite{Van2012A}. Therefore, the navigation mesh methods provide more flexibility than roadmap-based methods. 
	The Generalized Voronoi Diagram (GVD) is often used to decompose the free space into regions to construct a navigation mesh\cite{Geraerts2008Enhancing}.
	
	The navigation mesh is mainly designed for 2D polygonal environments \cite{Toll2011Navigation}, in which geometric regions can be extracted to represent free space effectively. In contrast, a map scanned from a real environment consists of discrete points, which represent obstacles with irregular shapes or noise. Hence, navigation mesh is not quite suitable to this type of maps. However, the Area Graph is entirely designed for the robot map from the real environment. In addition to constructing a set of one-dimensional edges for path planning based on the Area Graph, since the regions in the Area Graph are connected, the Area Graph  can also be developed as a navigation mesh in a real environment for the robots to make more flexible path planning.

	
	
	%

	\subsection{Passage Graph}
	\label{subsec:generation_passagegraph}

	The generation of the Passage Graph, as a roadmap based on the Area Graph, includes two parts. One is the off-line paths construction stage, another is the on-line query stage. 
	In the off-line path construction stage, the Passage Graph is generated for all areas and the roadmap paths between passages are stored. 
	In the on-line query stage, two points are given as a start and a goal. 
	First, the local Passage Graph for each given point is created by building paths to connect the given point, as a virtual passage, with the passages in the area that it locates in respectively.
	Then the path to connect the two given point is found by running A$ ^* $ search algorithm based on the vertices in the roadmap, passages in the Passage Graph. 
	
	The Passage Graph strongly depends on previous representation. It is denoted as  $G_p=(E_p,V_p,A_p)$. Here $A_p$ is the set of area cells. Each $v \in V_p$ is a passage vertex, which is the Voronoi vertices on the touched line of two neighboring areas. An edge $e\in E$ connects two passage vertices, which is attributed with a walkable path from one passage to another.
	Fig. \ref{fig:passageGraph} shows the Passage Graph for a map, in which only the edges between passage vertices are shown without the walkable paths attributed to them.

	Areas that have been merged after room detection, such as those in Fig. \ref{fig:dectRoomLab}, often have more than two passages. 
	To build a Passage Graph $ G_p $ from $ G_A $, an important task is to find a path between each  pair of passages.  
	We implemented two methods to build the paths in the Passage Graph.
	The first implementation simply runs A$ ^* $ planning algorithm based on grid map between passages for each room.
	In the second method, we employ the $A^*$ search \cite{hart1968a} on the Topology Graph to find paths between passages for each room, 
	then save these passage-passage paths. Because the paths are constructed not based on pixels, it saves time compared to building roadmap paths from a grid map. 
	
	In the on-line query step, we are given two points as start and goal. 
	We first check  which areas the two points are in. 
	For each given point, a new and virtual passage vertex is build and is added to the area where it located in.
	Then the virtual passage vertices will be connected to all passages of that area by adding corresponding edges between them.
	To generate the paths for those new edges, there are some differences for the two implementations. 	For A*-based Passage Graph implementation, we first check whether the two given points are in the same room. If they are, then we run A* planning between them in the room directly. Otherwise, the A* algorithm based on the grid map is run from the given point to each passage of the area that the point resides in. For the Voronoi-based methods, we connect a new virtual passage with the closest Topology Graph path waypoints with a straight line. 
	The properties of the Voronoi Diagram make sure that there cannot be any obstacle between the point and that graph. Then the given points are connected to the passages in the Passage Graph.
	
	Once we have connected the new virtual passage to the passages of the Passage Graph we can employ the $A^*$ search to find  the complete path from the start to the goal.
	
	\subsection{Results Analysis}
	\label{subsec:experiments}

	\begin{table}[t]
		\caption{Compare the different planning methods
		 \label{table:comparasonAstar3}}
		\begin{center}
			\begin{tabular}{|c|c|c||c|c|c||c|}
				\hline
				\multicolumn{3}{|c||}{Path Distance (m)} & \multicolumn{3}{|c||}{Planning Time (ms)} & Remark  \\
				\hline
				Grid & A*-P & Voro-P & Grid & A*-P & Voro-P & Cross\\
				\hline
				16.2 	& 16.2 	& 22.2 	& 444 	& 49 	& 9.0 	& Middle\\
				54.5 	& 54.5 	& 67.5 	& 446 	& 95 	& 9.7 	& Big \\
				26.2 	& 27.0 	& 34.8 	& 644 	& 192	& 8.5 	&  4 \\
				44.3 	& 47.8 		& 49.9 	& 532 	& 95 	& 5.5 	&  7 \\
				119.8 	& 125.3 	& 135.4 	& 1508 	& 428 	& 50.7 	&  11 \\
				95.7 	& 104.7 	& 110.9 	& 4841 	& 139 	& 6.5 	&  17 \\
				209.7 	& 221.7 	& 234.2 	& 5450 	& 176 	& 14.6 	&  23 \\
				\hline
			\end{tabular}
		\end{center}
	\end{table}
	
\begin{figure}[h!]
			\includegraphics[width=0.3\linewidth]{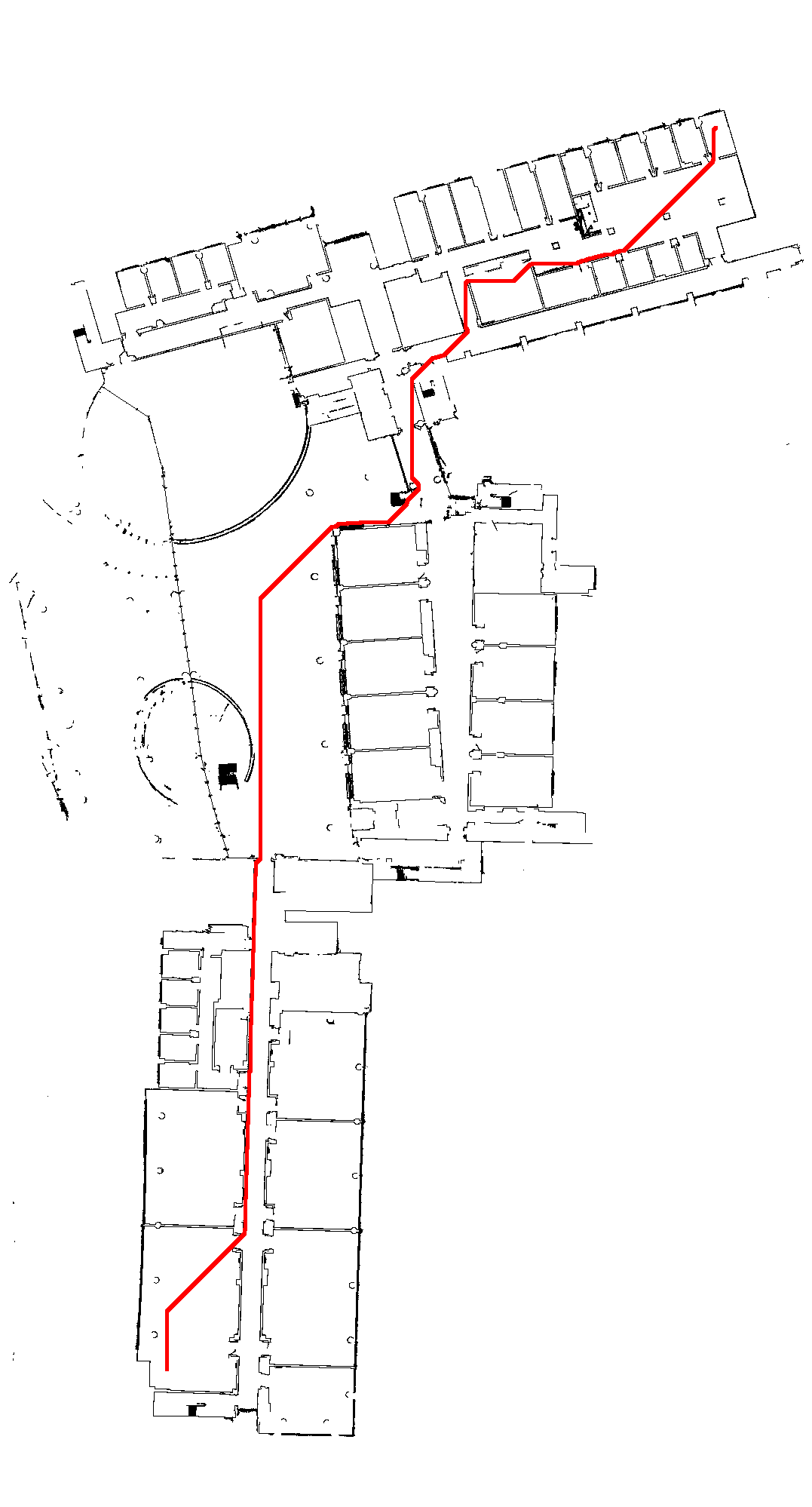}
			\includegraphics[width=0.3\linewidth]{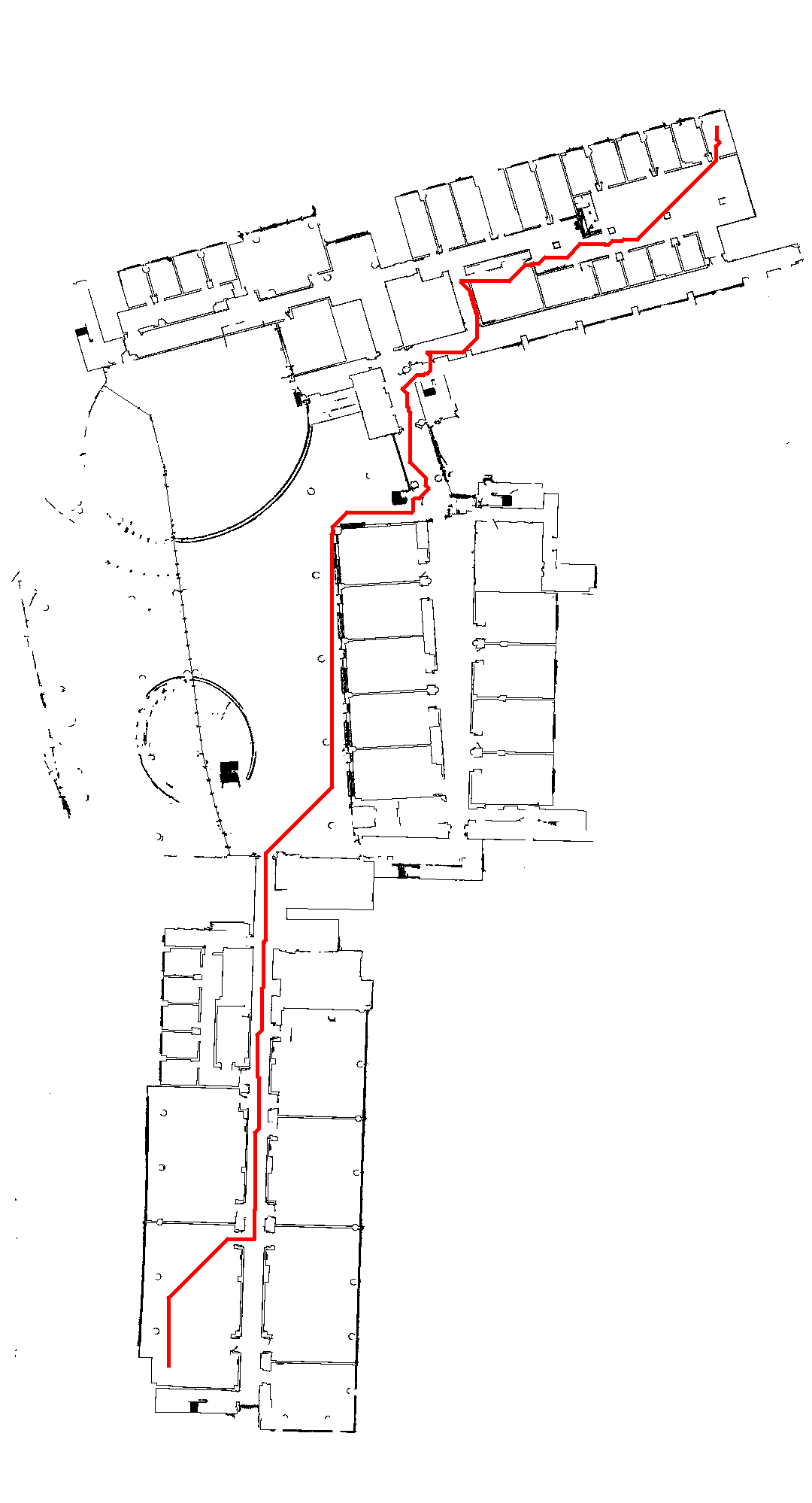}
			\includegraphics[width=0.3\linewidth]{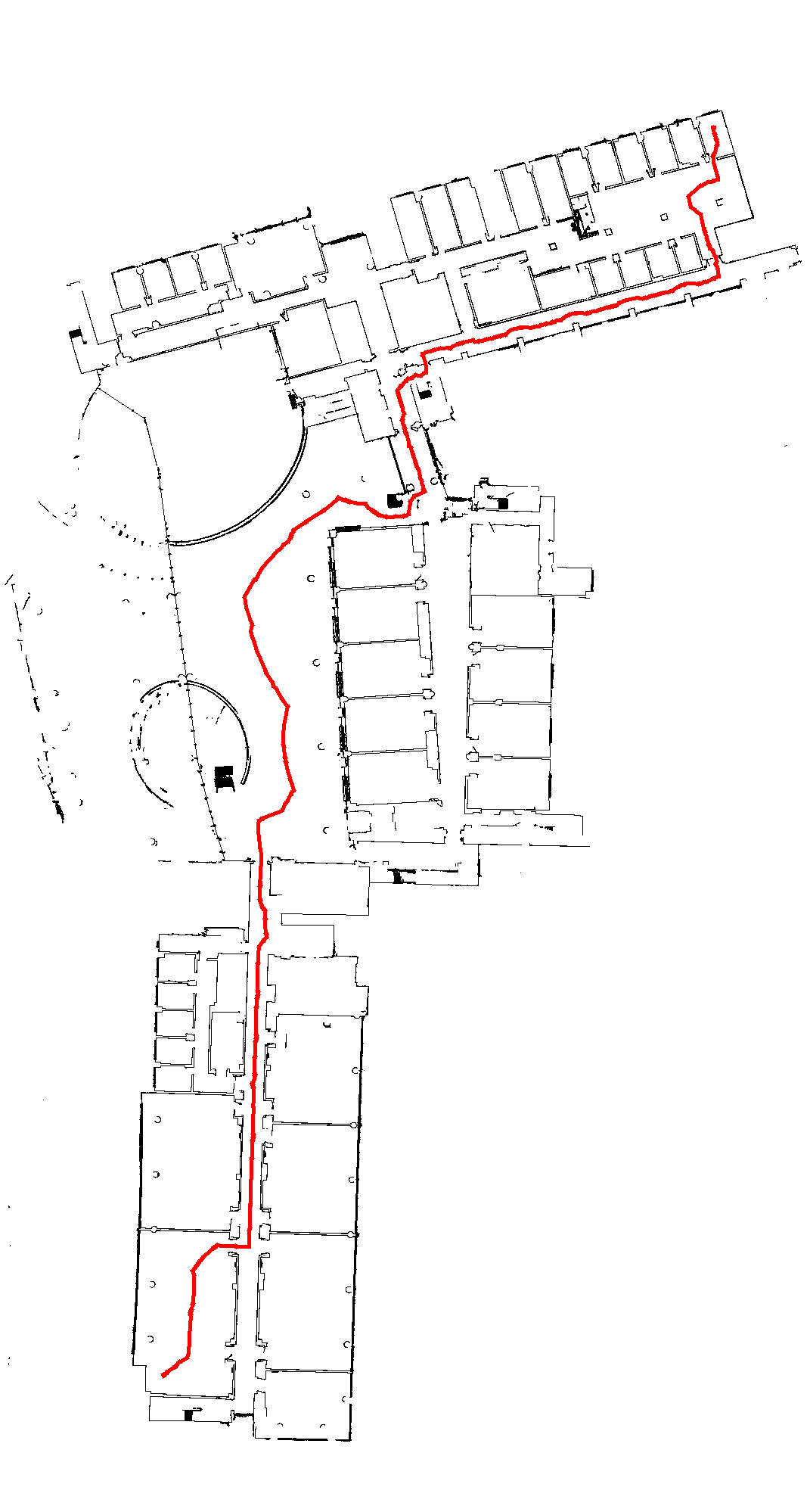}
		\caption{The planned paths obtained by A* on metric map (left), A*-based Passage Graph and Voronoi-based Passage Graph. \label{fig:pathresults}}
	\end{figure}

	Here we compare our algorithm with an A* implementation on the grid map.
    For that we use a big map of our building, which is edited   to remove some noise and add some walls that were missing.
	This map is has $ 2000 \times 1500 $ cells, set to 0.1 $ m $ per grid cell. The experiments were done on a Intel i7-4712HQ CPU @ 2.30GHz.
	The results are shown in Fig. \ref{fig:pathresults} and Table \ref{table:comparasonAstar3}. There ``Grid'' column is for A* on the grid map. ``A*-P'' and ``Voro-P'' mean that the results are obtained by running planning under A*-based Passage Graph and Voronoi-based Passage Graph, respectively. 
	
	The initialization time for the extraction of the Area Graph from the grid occupancy map is 26.0 seconds. We use the same Area Graph to construct an A*-based Passage Graph and a Voronoi-based Passage. The time for Voronoi-based Passage Graph extraction is 45ms, while the one for A*-based Passage Graph extraction is 20.4s. 


	The search complexity of the A* for grid map depends on the number of pixels, which is more than $ 10^6 $ in this map; the one for a Voronoi-based Passage Graph mainly depends on the number of areas and the number of  the passages in the room(s) that the given points located in, where the number of  rooms is only 302 even in this map; while the one for the A*-based Passage Graph includes the number of rooms and the pixels number and passages number in the room where the given point located in. 
	For Voronoi-based Passage Graph, it only takes longer time on planning in the fifth test. This is because in this case, the given point belongs to the biggest room that has 6 passages, which makes it take time to obtain paths to connect the given points with the passages. For the A*-based Passage Graph, it takes more time when a given point is in a big room because A* planning will be run on pixel level from the point to every passage.

	
	Observing the last three rows in Table \ref{table:comparasonAstar3}, we found that when the planning paths have to cross many rooms, the planning time of the A* algorithms is growing rapidly while that of the Passage Graph approach has very little growth, which shows that the advantage of the Area Graph-based planning will be even more obvious for even bigger maps. 
	

	\section{CONCLUSIONS}
	\label{sec:conclusions}
	
	
	In this paper, we proposed the Area Graph, that represents a map as a set of areas, and implemented the path planning as an application based on this representation. 
	We did an experiment considering the set of areas of the Area Graph as a segmentation of the map, and showed  that the proposed Area Graph construction method is more effective in avoiding the over-segmentation.
	
	It is worth noting that the role of the Area Graph goes far beyond planning and segmentation. 
	We are working on applying the area graph to other applications like navigation mesh, map merging or map evaluation, similar to the work in \cite{Schwertfeger2016PathMatching}. 
	Further work will investigate hierarchical Area Graphs. For example, after generated areas for a building, we can take the whole building as one area in the second level graph, then take a block with lots of buildings as an area in the third level graph. We are also exploring how to add more 3D information to the Area Graph, e.g. how to represent volumes instead of just areas.


	\addtolength{\textheight}{-11cm}   

	
	

	

	
	
	\bibliographystyle{IEEEtran}
	
	\bibliography{references}

\end{document}